\documentclass{article} 
\usepackage{iclr2026_conference,times}

\usepackage{csquotes}

\usepackage{amsmath,amsfonts,bm}

\def\eqref#1{equation~\ref{#1}}

\def\1{\bm{1}}

\DeclareMathAlphabet{\mathsfit}{\encodingdefault}{\sfdefault}{m}{sl}
\SetMathAlphabet{\mathsfit}{bold}{\encodingdefault}{\sfdefault}{bx}{n}

\usepackage[colorlinks,
            linkcolor=magenta,
            anchorcolor=blue,
            citecolor=gray
            ]{hyperref}

\usepackage{url}
\usepackage{adjustbox}
\usepackage{amsmath}
\usepackage{graphicx}
\usepackage{booktabs}
\usepackage{wrapfig}
\usepackage{xcolor}  
\usepackage{xcolor}
\usepackage{pifont}
\usepackage{soul}
\usepackage{url}
\usepackage[T1]{fontenc}
\usepackage{siunitx}
\usepackage{graphicx}
\usepackage{float}
\usepackage{multirow}
\usepackage{color}
\usepackage{wrapfig}
\usepackage{booktabs}
\usepackage{amssymb}
\usepackage{amsmath}
\usepackage{amsfonts}
\usepackage{cases}
\usepackage[utf8]{inputenc}
\usepackage{algorithm}
\usepackage{algpseudocode}
\usepackage{marvosym}
\usepackage{subfigure}
\usepackage{listings}
\usepackage{makecell}
\usepackage{amssymb,mathrsfs,amsmath}
\usepackage{amsmath, bm}
\usepackage{colortbl}
\usepackage[utf8]{inputenc}
\usepackage[T1]{fontenc}
\usepackage{wrapfig}
\usepackage{xcolor}
\usepackage{pdfpages}
\usepackage{tikz} 
\usepackage[most]{tcolorbox} 
\usepackage{lipsum}          
\definecolor{warningcolor}{RGB}{255, 0, 0}

\title{DeepScientist: Advancing Frontier-Pushing Scientific Findings Progressively}

\author{
  Yixuan Weng*, Minjun Zhu*, Qiujie Xie, Qiyao Sun, Zhen Lin, Sifan Liu, Yue Zhang\Letter \\
  Engineering School, Westlake University \\
  \texttt{wengsyx@gmail.com; \{zhu.minjun,zhangyue\}@westlake.edu.cn} \\[0.3cm]  
  \textbf{Project}: \url{https://ai-researcher.net} \\
  \textbf{Code}: \url{https://github.com/ResearAI/DeepScientist}
}

\iclrfinalcopy

\definecolor{OliveGreen}{RGB}{0,128,0}
\newcommand{\xqj}[1]{#1}

\begin{document}

\maketitle
\begin{abstract}

While previous AI Scientist systems can generate novel findings, they often lack the focus to produce scientifically valuable contributions that address pressing human-defined challenges. We introduce DeepScientist, a system designed to overcome this by conducting goal-oriented, fully autonomous scientific discovery over month-long timelines. It formalizes discovery as a Bayesian Optimization problem, operationalized through a hierarchical evaluation process consisting of "hypothesize, verify, and analyze". Leveraging a cumulative Findings Memory, this loop intelligently balances the exploration of novel hypotheses with exploitation, selectively promoting the most promising findings to higher-fidelity levels of validation. Consuming over 20,000 GPU hours, the system generated about 5,000 unique scientific ideas and experimentally validated approximately 1100 of them, ultimately surpassing human-designed state-of-the-art (SOTA) methods on three frontier AI tasks by 183.7\%, 1.9\%, and 7.9\%. This work provides the first large-scale evidence of an AI achieving discoveries that progressively surpass human SOTA on scientific tasks, producing valuable findings that genuinely push the frontier of scientific discovery.

\begin{figure}[!ht]
    \centering
    \includegraphics[width=\linewidth]
    {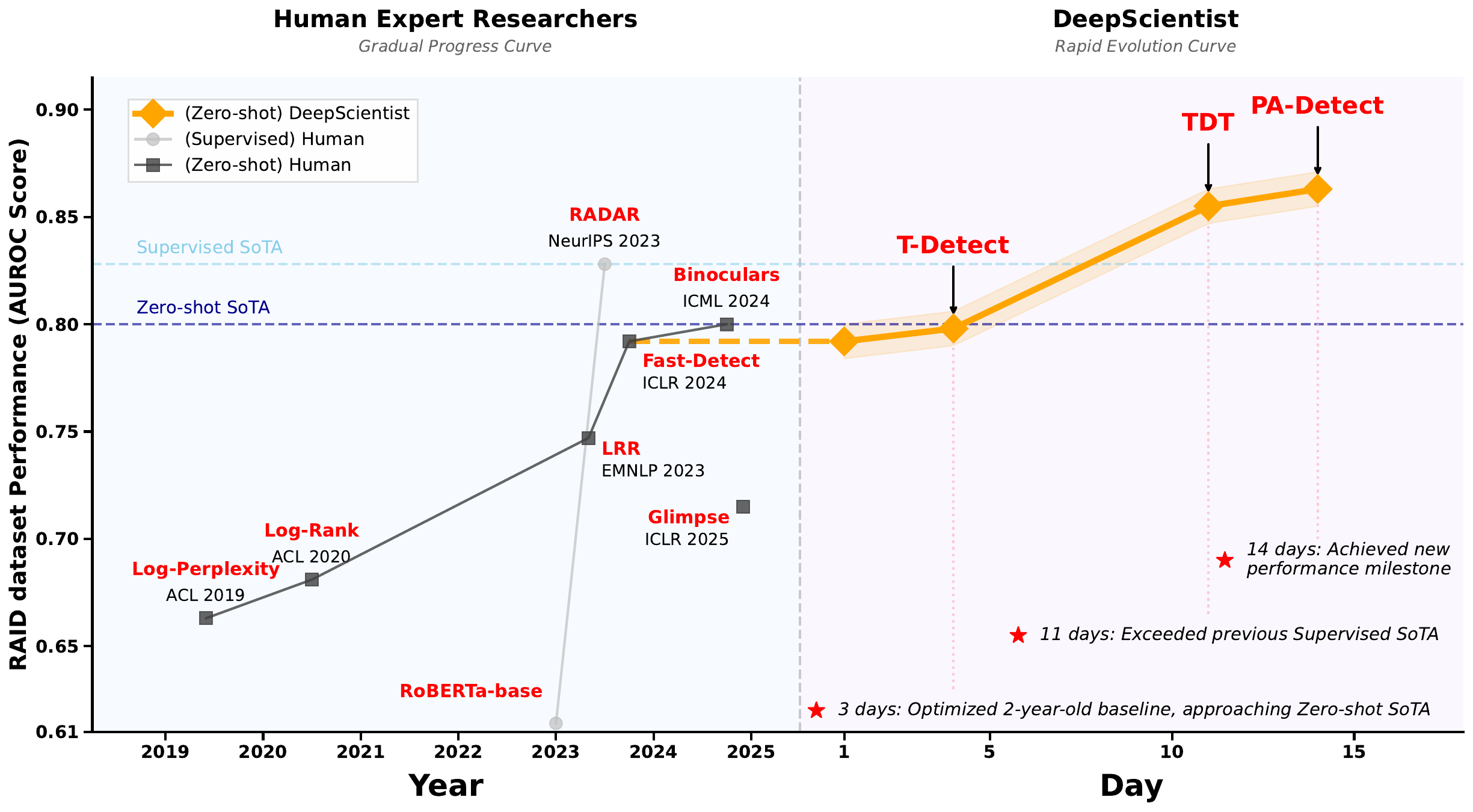}
    \vspace{-0.7cm}
    \caption{Comparison of research progress timelines for AI text detection on the RAID \citep{dugan-etal-2024-raid}. The right panel shows that DeepScientist achieves progress in two weeks that is comparable to three years of human research \citep{su2023detectllm,baoglimpse,baofast,hu2023radar} (left panel). All zero-shot methods, including the system-generated T-Detect, TDT, and PA-Detect, uniformly adopt Falcon-7B \citep{almazrouei2023falcon} as the base model. Additionally, all methods produced by DeepScientist demonstrate higher throughput than the previous SOTA method, Binoculars \citep{hans2024spotting}.}
    \label{fig:timelines}
    \vspace{-0.1cm}
\end{figure}
\end{abstract}

\section{Introduction}

Scientific discovery is inherently a process of \textbf{continuous exploration} and \textbf{trial-and-error}, where vast amounts of time and effort are invested to push the boundaries of human knowledge forward by a small step. This principle of persistent, incremental advancement is visible across the history of technology. For example, the decades-long optimization of semiconductor manufacturing has seen the feature size of transistors systematically reduced from micrometers to single-digit nanometers~\citep{moore1965moore}. Similarly, the efficiency of photovoltaic cells has been continuously advanced over half a century, with myriad material and architectural innovations pushing conversion rates from nascent single-digit percentages ever closer to their theoretical limits \citep{green1993silicon}. These historical trajectories underscore a process where human scientists engage in decades of \textbf{goal-directed}, \textbf{iterative} work to advance the SoTA artifacts continuously.

Recently, the emergence of Large Language Models (LLMs) has propelled \textbf{automated scientific discovery}, where LLM-based AI Scientist systems take the lead in exploration~\citep{xie2025far}.
With their powerful capacity for long-form generation and comprehension, LLMs enable \textbf{end-to-end, full-cycle automation in scientific discovery.} This has inspired influential work such as \textsc{AI Scientist-v2}~\citep{yamada2025ai}, whose scientific artifacts have been published in top-tier conference workshops. However, in the absence of clearly defined scientific goals, current AI Scientist systems often fall into the trap of blindly recombining existing knowledge and methods. As a result, their research outputs frequently appear naive under human evaluation and lack genuine scientific value~\citep{zhu2025ai}. \textbf{AI Scientists are yet to solve human challenges.}

To solve real-world challenges, We formally model the full cycle of scientific discovery as a \textbf{goal-driven Bayesian Optimization problem}, where the singular objective is to find a novel method that maximally improves a target performance metric. Building on this formulation, we introduce \textbf{DeepScientist}, a LLM-based agent system designed to explore progressively across the unknown space of possible candidate research methods to \textbf{identify the optimal plan} that maximizes a highly expensive-to-evaluate function of true scientific value. Specifically, DeepScientist employs an \textbf{iterative workflow}, together with a continuously expanding memory of prior research knowledge to efficiently manage uncertainty during exploration. It intelligently balances \textbf{exploitation}~(deepening investigations into promising high-value directions) with \textbf{exploration}~(venturing into uncharted areas to acquire new knowledge). Through large-scale parallel exploration, DeepScientist can generate innovative hypotheses and ultimately yield both valuable new methods and validation-proven scientific findings through continuous exploration.

We select three frontier scientific tasks~(\emph{Agent Failure Attribution}, \emph{LLM Inference Acceleration}, and \emph{AI Text Detection} ), take their state-of-the-art methods (\underline{ICML 2025 Spotlight}, \underline{ACL 2025 Outstanding}, \underline{ICLR 2024}) as starting points, and ask DeepScientist to conduct continuous research. As shown in Figures~\ref{fig:timelines} and \ref{fig:performance}, within a month-long cycle of exploration, validation, and iteration on 16 H800 GPUs, \textbf{DeepScientist exceeds their respective human SOTA methods by 183.7\% (Accuracy), 1.9\% (Tokens/second), and 7.9\% (AUROC) through autonomously redesigning core methodologies, rather than simply combining existing techniques} (Section~\ref{sec:achievements}). To understand how such progress emerged, we analyze DeepScientist’s discovery logs, and formed a small program committee to review the generated papers (Section~\ref{sec:Quality}). These logs show that the system generated over 5,000 unique ideas, of which only 1,100 are selected for experimental validation, and just 21 ultimately lead to scientific innovations (Section~\ref{sec:analysis}). Moreover, through the scaling experiment on computational resource, we discover a near-linear relationship between the resources allocated and the output of valuable scientific discoveries.

To our knowledge, we provide \textbf{the first empirical demonstration of an automated full-cycle scientific discovery system capable of producing novel, SOTA-surpassing methods and continuously advancing scientific frontiers} at a pace that substantially exceeds human researchers. Our findings reveal a stark reality: while the AI's exploratory speed is immense, its inherent success rate for innovation remains exceptionally low, making effective validation and filtering the new bottleneck at the frontier of automated science. Therefore, the central question of the field is no longer 'Can AI innovate?', but rather 'How can we efficiently guide its powerful, yet highly dissipative, exploratory process to maximize scientific return?' We hope this work can inspire the research community to develop AI Scientist systems with greater exploration efficiency to accelerate scientific discovery at a larger scale, paving the way for ground-breaking discoveries.

\section{Related Work}

\textbf{Replication and Optimization.} A significant body of research focuses on engineering tasks that operate within established scientific frameworks. This includes replication-oriented works like PaperBench \citep{starace2025paperbench} and Paper2Agent \citep{miao2025paper2agentreimaginingresearchpapers}, which aim to reproduce existing papers. Other works, such as Agent Laboratory \citep{schmidgall2025agent} and MLE-Bench \citep{chan2024mle}, tackle early-stage machine learning engineering problems. Similarly, systems like AlphaTensor \citep{fawzi2022discovering} and AlphaEvolve \citep{novikov2025alphaevolve} use massive trial-and-error with known engineering methods to improve the performance of codebases. The common goal of these efforts is engineering-driven optimization within an established scientific paradigm, enhancing existing systems without questioning their foundational assumptions. DeepScientist, in contrast, pursues scientific discovery by targeting the core limitations of the SOTA itself. Its objective is not to refine the current state-of-the-art, but to establish a new one by introducing fundamentally different methodologies.

\textbf{Semi-Automated Scientific Assistance.} The path toward automating scientific discovery begin not with replacing the scientist, but with assisting them, leading to the development of a paradigm of specialized AI tools for individual research tasks. Systems like CycleResearcher \citep{weng2025cycleresearcher} handle writing, DeepReview \citep{zhu2025deepreview} manages reviewing, and co-scientists \citep{gottweis2025towards,penades2025ai,swanson2025virtual,baek2025researchagent} aid in hypothesis generation. These powerful tools address only isolated fragments of the scientific process, leaving the crucial loop of learning \xqj{from failure and exploration to humans. In contrast, DeepScientist is an autonomous agent of inquiry, managing the entire end-to-end research cycle and closing the loop by learning from its own experiments and self-directing its research path.}

\xqj{\textbf{Automated Scientific Discovery.} Building on the capabilities of specialized assistants, a line of research pursue full, end-to-end research automation~\citep{yang2023ai,xie2025an}. Pioneering efforts, such as the AI Scientist systems \citep{lu2024ai, yamada2025ai} and subsequent work \citep{zochi2025,airesearcher}, successfully demonstrate that an AI system could manage the full research cycle and produce novel findings. However, their primary limitation often lies in their exploratory strategy, which lacks a specific scientific goal rooted in a field's grand challenges, resulting undirected discoveries that may be perceived as lacking genuine scientific value. Instead, DeepScientist is thus the first automated scientific discovery system that leverages a closed-loop, iterative process to discover methods surpassing the human state-of-the-art. The exploration of DeepScientist is goal-oriented and insight-driven, beginning by identifying a recognized limitation in the human SOTA and then using failure attribution to ensure discoveries are both novel and scientifically meaningful.}

\section{DeepScientist: A Progressive System for Discovering SOTA-Surpassing Findings}

\begin{figure}[h]
    \centering
    \includegraphics[width=\linewidth]{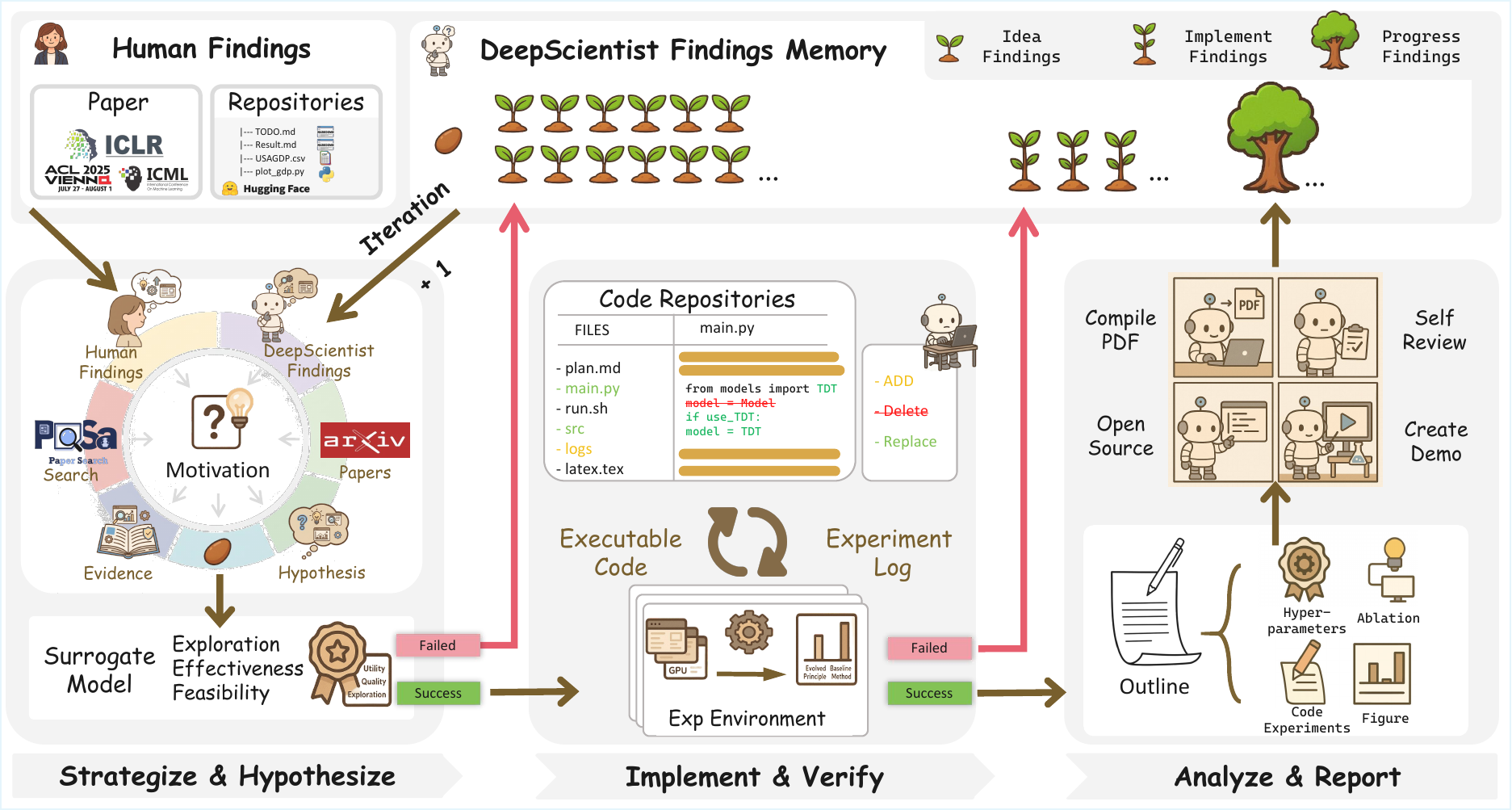}
    \vspace{-0.6cm}
    \caption{The autonomous, closed-loop discovery process of DeepScientist. The system iterates through a three-stage cycle, learning from both human knowledge and its own experiments. 
    }
    \label{fig:main_workflow}
\end{figure}

\subsection{Modeling Scientific Discovery as an Optimization Problem}
The fundamental goal of automated scientific discovery is to autonomously identify novel methods that yield significant advancements in a given scientific domain. This process can be formally conceptualized as a search for an optimal solution within a vast and unstructured space of possibilities. Let the space of all possible candidate research methods be denoted by $\mathcal{I}$. Each individual method $I \in \mathcal{I}$, such as a novel algorithm or a new model architecture, possesses an intrinsic scientific value. This value is determined by a latent, black-box true value function, $f: \mathcal{I} \to \mathbb{R}$, which maps a method to its ultimate empirical impact. \textbf{The objective of scientific discovery is therefore to find the optimal method $I^{*}$ that maximizes this function:}
\begin{equation} \label{eq:optimization_goal}
    I^{*} = \operatorname{arg\,max}_{I \in \mathcal{I}} f(I)
\end{equation}

Unlike previously studied tasks such as early-stage machine learning \citep{AgentLaboratoryUsingLLMAgents_from_json3}, algorithmic design \citep{novikov2025alphaevolve,lange2025shinkaevolve}, or scientific software development \citep{aygün2025aihelpscientistswrite}, \textbf{A defining characteristic of frontier scientific discovery is that each exploratory step demands immense computational and intellectual resources, making the evaluation of the true scientific value function, $f(\cdot)$, prohibitively costly.} Any single evaluation, $f(I)$, corresponds to a complete and resource-intensive research cycle of implementation, experimentation, and analysis, often consuming vast computational resources (e.g., on the order of $10^{16}$ FLOPs for a frontier LLM problem, as illustrated in Figure~\ref{fig:placeholder1}.c). This extreme sample inefficiency renders brute-force or random exploration of the space $\mathcal{I}$ intractable. Therefore, we model the problem within the framework of Bayesian Optimization\citep{frazier2018tutorial,garnett2023bayesian}, which provides a principled methodology for global optimization of expensive black-box functions. By constructing a surrogate model to intelligently guide the search, Bayesian Optimization effectively reduces the number of costly real-world evaluations through a careful balance of exploration and exploitation. However, 
\textbf{for scientific discovery, $\mathcal{I}$ is a conceptual space that is not explicitly defined.} Candidate methods $I$
must be formulated as creative, plausible, and coherent scientific hypotheses. The generation of high-quality candidate hypotheses is a critical bottleneck that traditional Bayesian Optimization algorithms are not designed to address. This challenge necessitates a new mechanism that integrates creative ideation with sample-efficient optimization. We detail our solution to this problem in the following subsections.

\subsection{The DeepScientist Framework}

The architecture of DeepScientist actualizes the Bayesian Optimization loop through a multi-agent system equipped with an \textbf{open-knowledge system} and a continuously accumulating \textbf{Findings Memory}. This memory is composed of both frontier human knowledge (e.g., papers and codes) and the system's own historical findings, and it intelligently guides subsequent explorations. \textbf{The entire discovery process is structured as a hierarchical and iterative three-stage exploration cycle}. In this hierarchical scheme, only research ideas that exhibit promise are advanced to more expensive evaluations, while others are retained in the Findings Memory to inform subsequent explorations. This design ensures the computational resources are dynamically and precisely allocated to the most promising scientific trajectories, thereby \textbf{maximizing discovery efficiency under constrained budgets}. Specifically, each stage within the three-stage exploration cycle is associated with a distinct \textbf{fidelity–cost tradeoff}~(Figure \ref{fig:main_workflow}):

\textbf{Strategize \& Hypothesize.}
Each research cycle begins by analyzing the Findings Memory ($\mathcal{M}_t$), a list-style database containing thousands of structured records. Each record represents a unique scientific finding, which is categorized according to its stage of development. To overcome the LLM's context length constraints, we use a separate retrieval model \citep{wolters2024memoryneedoverviewcomputeinmemory} when needed to select the Top-K Findings as input. The vast majority of records begin as Idea Findings—unverified hypotheses. During this first stage, the system identifies limitations in existing knowledge and generates a new collection of hypotheses ($\mathcal{P}_{\text{new}}$), and then they evaluated by a low-cost Surrogate Model ($g_t$). The surrogate model (an LLM Reviewer) is first contextualized with the entire Findings Memory. It then approximates the true value function $f$ and, for each candidate finding $I \in \mathcal{P}_{\text{new}}$, produces a structured valuation vector $V = \langle v_u, v_q, v_e \rangle$, quantifying its estimated utility, quality, and exploration value as integer scores on a scale of 0 to 100. Each new hypothesis and its valuation vector is then used to initialize a new record in the Findings Memory as an "Idea Finding".

\textbf{Implement \& Verify.}
\xqj{This stage serves as the primary filter in the Findings Memory.} To decide which of the numerous "Idea Findings" warrants the significant resource investment to be advanced in a real-world experiment, the system employs an Acquisition Function ($\alpha$). Specifically, it uses the classic Upper Confidence Bound (UCB) algorithm to select the most promising record. The UCB formula maps the valuation vector $V$ to balance the trade-off between exploiting promising avenues (represented by $v_u$ and $v_q$) and exploring uncertain ones (represented by $v_e$):
\begin{equation} \label{eq:bo_cycle}
 I_{t+1} = \operatorname{arg\,max}_{I \in \mathcal{P}_{\text{new}}} \big( \underbrace{w_u v_u + w_q v_q}_{\text{Exploitation Score}} + \kappa \cdot \underbrace{v_e}_{\text{Exploration Score}} \big),
\end{equation}
where $w_u$ and $w_q$ are hyperparameters and $\kappa$ controls the intensity of exploration. The highest-scoring finding $I_{t+1}$ is selected for validation, and its record is promoted to the status of an Implement Finding. \textbf{A coding agent then performs a repository-level implementation to executed the experiment.} This agent operates within a sandboxed environment with full permissions, allowing it to read the complete code repository and access the internet for literature and code searches. Its objective is to implement the new hypothesis on top of the existing SOTA method's repositories. The agent typically begins by planning the task, then reads the code to understand its structure, and finally implements the changes to produce the experimental logs and results. The experiment logs and results, $f(I_{t+1})$, is used to update the corresponding record, enriching it with empirical evidence and thus closing the learning loop.

\textbf{Analyze \& Report.}
\xqj{The final and most selective stage of the Findings Memory is triggered only by a successful validation.} When an "Implement Finding" succeeds in surpassing the baseline, its record is promoted to a Progress Finding. \xqj{This transformation is implemented by a series of specialized agents capable of utilizing a suite of MCP \citep{hou2025model} tools.} These agents first autonomously design and execute a series of deeper analytical experiments (e.g., ablations, evaluations on new datasets), leveraging MCP tools to manage the experimental lifecycle, data collection, and result parsing. Subsequently, a synthesis agent employs the same toolset to collate all experimental results, analytical insights, and generated artifacts into a coherent, reproducible research paper. \xqj{This deeply validated record becomes a new record in the system's knowledge base, thus influencing the decision-making process in all subsequent cycles.}

\begin{table}
    \centering
    \caption{Overview of the three different human SOTA methods we selected.}
    \label{tab:sota_baselines_en}
    \vspace{5pt}
    \resizebox{\textwidth}{!}{
    \begin{tabular}{lllll}
        \toprule
        \textbf{Task} & \textbf{Method} & \textbf{Venue} & \textbf{Benchmark} &\textbf{Github Star}\\
        \midrule
        Agents Failure Attribution & All at Once & ICML 2025 Spotlight & Who\&When& 302 \\
        \rowcolor[rgb]{ .949,  .949,  .949}
        LLM Inference Accel. & TokenRecycling & ACL 2025 Outstanding & MBPP& 323 \\
        AI Text Detection & FastDetectGPT & ICLR 2024 & RAID& 414 \\
        \bottomrule
    \end{tabular}}
    \vspace{-15pt}
\end{table}

\section{Experiments}

\xqj{As detailed in Table \ref{tab:sota_baselines_en}, we select three distinct SOTA methods (published in 2024 and 2025) as starting points, chosen for their frontier status, community interest, and human supervisability.} Each SOTA method is manually reproduced, and we preserve execution logs and test scripts to allow DeepScientist to focus on research advancement. DeepScientist is provided with two servers, each with 8 Nvidia H800 GPUs. To maximize utilization, we launch a separate system instance for each GPU, employing the Gemini-2.5-Pro model for core logic and the Claude-4-Opus model for its robust code-generation capabilities. Three human experts supervise the process to verify outputs and filter out hallucinations. \xqj{For more implementation details, please see Appendix \ref{appendix:impl}.}

\subsection{DeepScientist achievements on three research domains}
\label{sec:achievements}

\begin{figure}[h]
    \centering
    
    \adjustbox{width=\textwidth}{%
        \begin{tabular}{l|ll|l|ll}
        \toprule
        \multirow{2}{*}{\textbf{Method}} & \multicolumn{2}{c|}{\textbf{Agent Failure Attribution}} & \textbf{LLM Inference Acceleration} & \multicolumn{2}{c}{\textbf{AI Text Detection}} \\
        & Handcraft (Acc.) & Algorithm-Gen (Acc.) & Tokens/second & AUROC & Latency \\
        \midrule
        Human SoTA method & 12.07\% (All at Once) & 16.67\% (All at Once) & 190.25 (Token Recycling) & 0.800 (Binoculars) & 117ms (Binoculars) \\
        DeepScientist's method & \textbf{29.31\%} (A2P) & \textbf{47.46\%} (A2P) & \textbf{193.90} (ACRA) & \textbf{0.863} (PA-Detect) & \textbf{60ms} (PA-Detect) \\
        \midrule
        \textbf{Improvement} & \textcolor{red}{\textbf{$\Delta$+142.8\% (+17.24)}} & \textcolor{red}{\textbf{$\Delta$+183.7\% (+30.79)}} & \textcolor{red}{\textbf{$\Delta$+1.9\% (+3.65)}} & \textcolor{red}{\textbf{$\Delta$+7.9\% (+0.063)}} & \textcolor{red}{\textbf{$\Delta$+190\%$\downarrow$ (-57)}} \\
        \bottomrule
        \end{tabular}%
    }

    \includegraphics[width=\linewidth]{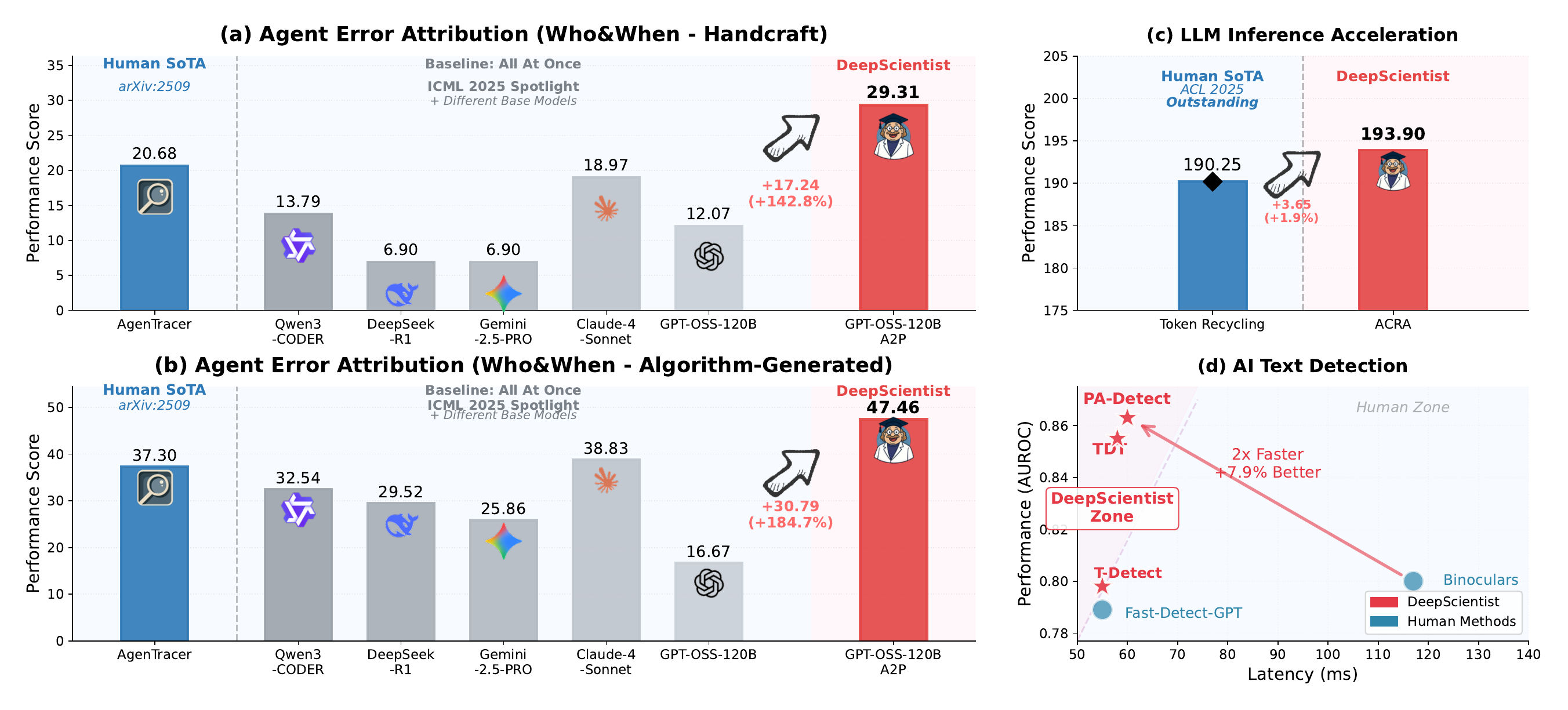}
    
    \vspace{-0.4cm}
    
    \caption{Performance evaluation of DeepScientist across three research domains: (a-b) Agent Failure Attribution on Who\&When benchmark in handcraft and algorithm-generated settings; (c) LLM Inference Acceleration on MBPP dataset; (d) AI Text Detection with performance-latency tradeoff analysis. DeepScientist (shown in pink) consistently outperform human-designed SoTA approaches (shown in blue) across all tasks.}
    \label{fig:performance}
\end{figure}

\textbf{Agents Failure Attribution.} The task addresses the question: in an LLM-based multi-agent system, which agent caused the task to fail and when? Starting from the baseline "All at once" method \citep{zhang2025which}, DeepScientist identified that the current approach lacks the counterfactual reasoning capabilities essential for attribution. Through a process of trial, error, and synthesizing new findings—discovering the effectiveness of hypothetical prediction and simulated attempts—it ultimately proposed the \textit{\underline{A2P}} method. Named for its Abduction-Action-Prediction process, its core innovation elevates failure attribution from pattern recognition to causal reasoning, filling the critical gap in counterfactual capabilities by predicting if a proposed fix would have led to success. As shown in Figure \ref{fig:performance}.(a-b), A2P achieved scores of 29.31 and 47.46 in the "handcraft" and "algorithm-generated" settings of the Who\&When benchmark, respectively, setting a new state-of-the-art (SOTA). In this task, DeepScientist validated that a structured, zero-shot causal reasoning framework can be superior to less principled methods. As of September 2025, the training-free A2P method maintains its SOTA position, outperforming even 7B models trained on synthetic data. \citep{zhang2025agentracerinducingfailurellm}.

\textbf{LLM Inference Acceleration} is a highly optimized field aiming to maximize throughput and reduce latency during LLM inference \citep{xia-etal-2024-unlocking}. In this process, the system actively made many different attempts, such as using a Kalman Filter \citep{zarchan2005progress} to dynamically adjust an adjacency matrix to address the original method's lack of a memory function. Although most of these attempts failed, the system-generated \textit{\underline{ACRA}} method ultimately advanced the MPBB \citep{austin2021program} from a human SOTA of 190.25 to 193.90 tokens/second by identifying stable suffix patterns, as shown in Figure \ref{fig:performance}. Scientifically, this innovation is significant because it uses this extra contextual information to dynamically adjust the decoding guess, effectively grafting a long-term memory onto the process and breaking the context-collapsing of standard decoders. This discovery highlights the system's primary goal: the creation of new, human-unknown knowledge rather than mere engineering optimization. For instance, one could likely achieve greater performance gains by combining ACRA with an established technique like layer skipping \citep{wang2022skipbert} or PageAttention \citep{kwon2023efficient}, but this would represent an engineering effort, not a scientific one. The exploration assessment within our process avoids such combinations of existing knowledge.

\textbf{AI Text Detection}  is a binary classification task where, given a text that may contain content from an LLM (and possibly additional noise), the goal is to determine if it was produced by a human or an LLM~\citep{li2022artificial,ghosal2023a}. To validate its capacity for sustained advancement, DeepScientist made numerous attempts that included addressing the Boundary-Aware Extension problem and exploring approaches like Volatility-Aware and Wavelet Subspace Energy methods. The final results show a dramatic acceleration in scientific discovery: in a rapid evolution over just two weeks, the system produced three distinct, progressively superior methods (\textit{\underline{T-Detect}}, \textit{\underline{TDT}}, and \textit{\underline{PA-Detect}}). This began with T-Detect fixing core statistics with a robust t-distribution, then evolved conceptually with TDT and PA-Detect, which treat text as a signal and use wavelet and phase congruency analysis to pinpoint anomalies. Scientifically, this shift reveals the "non-stationarity" of AI-generated text, alleviating the information bottleneck in prior paradigms that average away localized evidence. As shown in Figure \ref{fig:timelines} and \ref{fig:performance}(d), this entire discovery trajectory demonstrates DeepScientist's ability for advancing frontier-pushing scientific findings progressively, establishing a new SOTA with a 7.9\% higher AUROC while also doubling the inference speed.

\subsection{Assessing the Quality of AI-Generated Research Paper}
\label{sec:Quality}

\begin{table}[t]
\centering
\small
\caption{
Evaluation of AI-generated papers produced by various AI Scientist systems. Scores represent the average ratings given by DeepReviewer-14B~\citep{zhu2025deepreview} across the number (``Num'') of available papers. Note: Publicly available papers may be curated and therefore may not fully represent the typical output of each system.
}
\label{tab:ai_scientist_performance_llm_judge}
\vspace{5pt}
\begin{adjustbox}{width=\textwidth}
\begin{tabular}{lcccccc}
\toprule
\textbf{AI Scientist Systems} & \textbf{Number}  & \textbf{Soundness} & \textbf{Presentation} & \textbf{Contribution} & \textbf{Rating}& \textbf{Accept Rate}\\
\midrule
\textsc{AI Scientist} & 10  & 2.08 & 1.80 & 1.75 & 3.35&0\%\\
\rowcolor[rgb]{ .949,  .949,  .949}
HKUSD AI Researcher & 7 & 1.75 & 1.46 & 1.57 & 2.57 &0\%\\
\textsc{AI Scientist-v2} & 3 & 1.67 & 1.50 & 1.50 & 2.33 &0\% \\
\rowcolor[rgb]{ .949,  .949,  .949}
CycleResearcher-12B & 6  & 2.25 & 1.75 & 2.13 & 3.75&0\%\\
Zochi & 2  & 2.38 & 2.38 & 2.25 & 4.63&0\% \\
\rowcolor[rgb]{ .902,  .834,  .767}
DeepScientist (Ours) & 5  & \textbf{2.90} & \textbf{2.90} & \textbf{2.90} & \textbf{5.90}&\textbf{60\%} \\
\bottomrule
\end{tabular}
\end{adjustbox}
\end{table}

\begin{table}[t]
\centering
\tiny
\caption{
Evaluation of DeepScientist's papers produced by human experts. 
Values are presented as mean (variance) from three reviewers. Inter-rater reliability for Rating: Krippendorff's $\alpha$ = 0.739.
}
\label{tab:ai_scientist_performance_human_judge}
\vspace{5pt}
\begin{adjustbox}{width=\textwidth}
\begin{tabular}{lllllll}
\toprule
\textbf{Paper}  & \textbf{Confidence}& \textbf{Soundness} & \textbf{Presentation} & \textbf{Contribution} & \textbf{Rating}\\
\midrule
HUMAN Avg. (ICLR 2025) & - & 2.59 & 2.36 &  2.62 & 5.08 \\
\midrule
\textsc{1. \texttt{T-Detect}} & 4.33 (0.33) & 2.00 (1.00) & \textbf{2.67} (0.33) & \textbf{2.67} (0.33) & 5.00 (0.00)\\
\rowcolor[rgb]{ .949,  .949,  .949}
\textsc{2. \texttt{TDT}} & 4.67 (0.33) & \textbf{3.00} (0.00) & \textbf{3.00} (0.00) & \textbf{3.00} (0.00) & \textbf{5.67} (0.33)\\
\textsc{3. \texttt{PA-Detect}} & 4.00 (0.00) & 1.67 (0.33) & 2.00 (1.00) & 2.00 (1.00) & 4.33 (1.33)\\
\rowcolor[rgb]{ .949,  .949,  .949}
\textsc{4. \texttt{A2P}} & 4.00 (0.00) & \textbf{3.00} (0.00) & \textbf{3.00} (0.00) & \textbf{2.67} (0.33) & \textbf{5.67} (0.33)\\
\textsc{5. \texttt{ACRA}} & 3.33 (0.33) & 1.67 (0.33) & 2.00 (1.00) & 1.67 (0.33) & 4.33 (1.33)\\
\rowcolor[rgb]{ .902,  .834,  .767}
DeepScientist Avg. & 4.07 & 2.27 & \textbf{2.53} & 2.40 & 5.00\\
\bottomrule
\end{tabular}
\end{adjustbox}
\end{table}

\textbf{Experimental Setup.}
To assess the quality of the final output, we evaluate the five research papers autonomously generated by DeepScientist's end-to-end process. Our evaluation protocol is twofold. First, to benchmark against existing work, we employ DeepReviewer~\citep{zhu2025deepreview}, an AI agent that simulates the human peer-review process with an external search capability, comparing DeepScientist's output against 28 publicly available papers from other AI Scientist systems. Second, for a more rigorous assessment, we convene a dedicated program committee consisting of three active LLM researchers: two volunteers who have served as ICLR reviewers and one senior volunteer who has been invited to be an ICLR Area Chair. The generated papers are available in Appendix \ref{appendix:impl}.

\textbf{Automated Review Against Other AI Scientist Systems.}
As shown in Table~\ref{tab:ai_scientist_performance_llm_judge}, the results from the LLM-based automatic evaluation indicate that the system's outputs are recognized for their scientific novelty and value. \xqj{When benchmarked against 28 publicly available papers from other AI Scientist systems using DeepReviewer, DeepScientist is the only AI Scientist system to produce papers that achieves a 60\% acceptance rate.}

\textbf{Human Expert Evaluation.}
The evaluation from our human program committee, shown in Table \ref{tab:ai_scientist_performance_human_judge}, reveal a remarkable and unanimous consensus: DeepScientist consistently excels at ideation, the most challenging and often rate-limiting step in human-led research. Full details on the review protocol are provided in Appendix \ref{appendix:review}, and the core ideas within each paper are praised for their genuine novelty, ingenuity, and scientific contributions. The quality of these innovations is further demonstrated by the review scores: the system's average rating (5.00) closely mirrors the average of all ICLR 2025 submissions (5.08), with two of its papers significantly exceeding this (5.67).

\subsection{Analysis of the Iterative Trajectory of Autonomous Exploration}
\label{sec:analysis}

\begin{figure}[t]
    \centering
    \includegraphics[width=\linewidth]{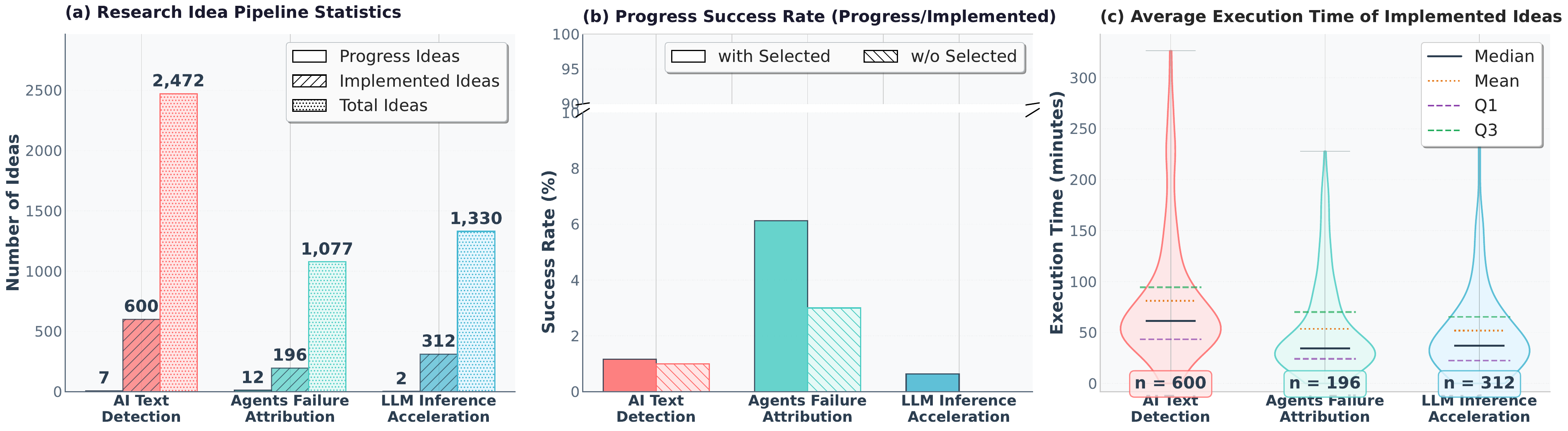}
    \vspace{-0.2cm}
    \caption{DeepScientist's experimental statistics. (a) The research pipeline from generated ideas to validated progress. (b) Success rates comparing our selection strategy against a baseline. (c) Distribution of wall-clock execution times for all implemented trials.}
    \label{fig:placeholder1}
\end{figure}
\begin{figure}[t]
    \centering
    \includegraphics[width=\linewidth]{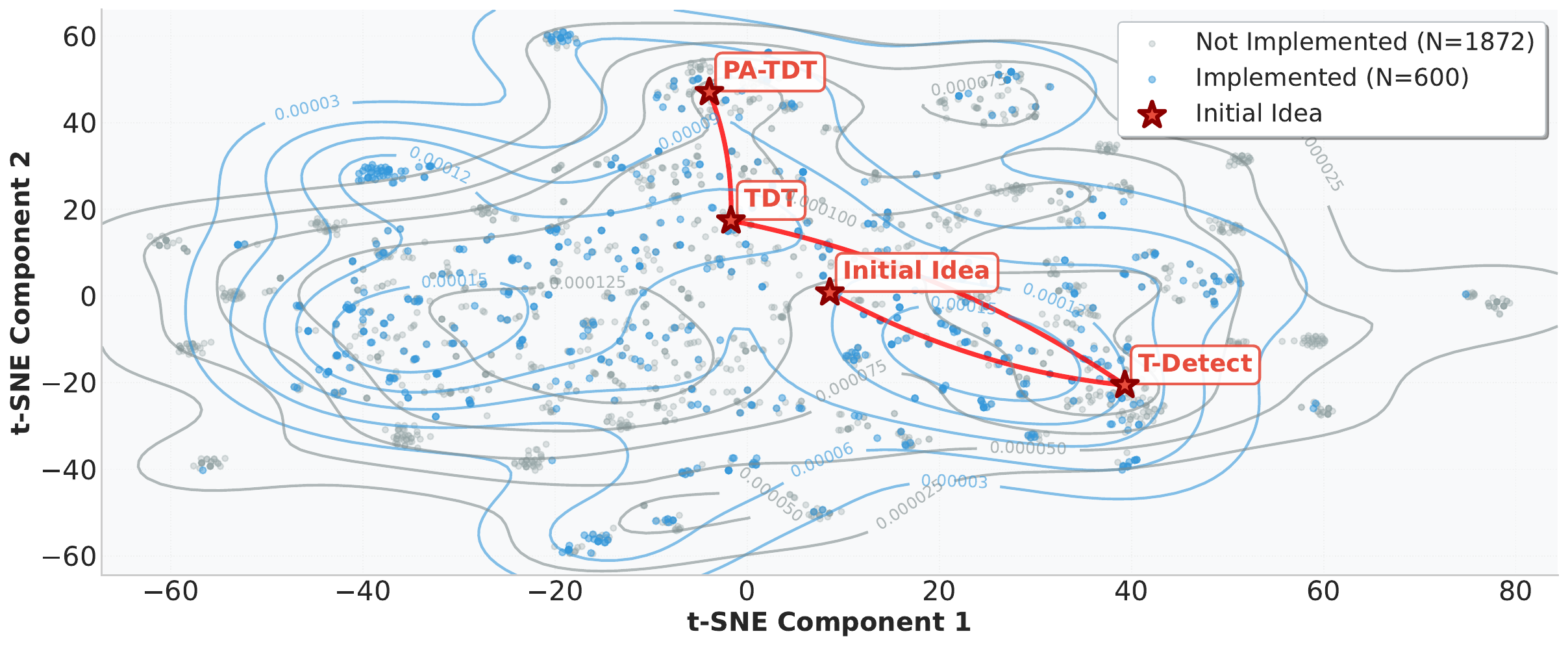}
    \vspace{-0.8cm}
    \caption{Visualization of the conceptual search space for the AI text detection task. The plot shows a t-SNE visualization of the semantic embeddings for all 2,472 generated ideas. Markers identify the initial SOTA method (Initial Idea) and the three final SOTA-surpassing methods (Progress Ideas).}
    \vspace{-0.2cm}
    \label{fig:placeholder2}
\end{figure}
\begin{figure}[t]
    \centering
    \includegraphics[width=\linewidth]{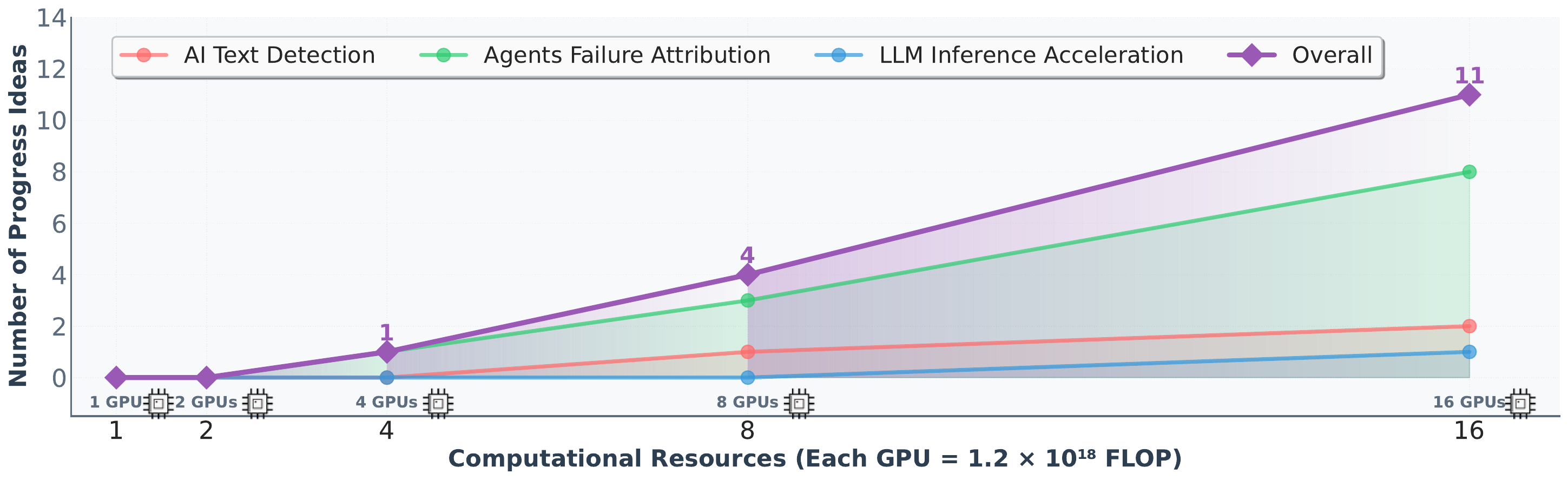}
    \vspace{-0.8cm}
    \caption{Scaling analysis of autonomous scientific discovery. The plot illustrates the relationship between parallel computational resources (number of GPUs) and the number of SOTA-surpassing "Progress Findings" found by DeepScientist across all tasks within a one-week period.}
    \vspace{-0.2cm}
    \label{fig:placeholder3}
\end{figure}

\textbf{Experimental Setup.}
The findings in this section are derived from a series of post-hoc analyses conducted on the complete operational data generated by DeepScientist across the three frontier tasks. This data includes the full set of execution logs and the Findings Memory, providing the basis for all subsequent statistical analysis. To visualize the conceptual search space (Figure \ref{fig:placeholder2}), we embed the complete description of each generated finding using the Qwen3-Embedding-8B model. To assess scalability (Figure \ref{fig:placeholder3}), we conduct a dedicated one-week experiment where N identified limitations of a single SOTA method are assigned to N parallel GPU instances. These instances explore solutions independently but share their findings to a central database, which are synchronized globally every five cycles to accommodate the asynchronous nature of the discovery process. Finally, to better understand the low success rate, our program committee experts perform a detailed causal attribution analysis on a sample of 300 failed implementations.

\textbf{Our analysis of DeepScientist's experimental logs reveals the sheer scale of the trial-and-error process inherent in autonomous scientific discovery.} Even in our relatively fast-executing domains, achieving progress required hundreds of trials per task. As show in Figure \ref{fig:placeholder1}, the execution time distributions show that while individual experiments may be quick, the sheer volume of trial-and-error necessary to uncover a successful idea is substantial. This suggests a clear application boundary for current autonomous science: for tasks with rapid feedback loops, such as knowledge editing or aspects of chip design, delegating massive-scale experimentation to AI is a powerful strategy. However, for high-cost endeavors like pre-training foundation models or pharmaceutical synthesis, the low success rate makes such an approach currently impractical, mandating continued reliance on human-led ideation. The autonomous research process is characterized by a vast exploratory funnel where promising ideas are exceptionally rare. Across the three tasks, DeepScientist generate over 5,000 unique ideas, yet only about 1,100 are deemed worthy of experimental validation by the system's selection mechanism, and a mere 21 ultimately result in scientific progress. An ablation study underscores the criticality of this selection process: without it, randomly sampling 100 ideas for each task and testing them yields a success rate of effectively zero. With our selection strategy, the success rate rises to approximately 1-3\%, demonstrating that while still low, intelligent filtering is essential. The low success rate is not merely a matter of failed hypotheses; analysis by human experts on a sample of failed trials reveals that approximately 60\% were terminated prematurely due to implementation errors, while the vast majority of the remaining 40\% simply offered no performance improvement or caused a regression. This highlights that the probability of an LLM-generated idea being both correct in its premise and flawless in its implementation is exceedingly low. The success of this work, therefore, is not a product of brute-force computation but of search efficiency. A naive approach of fully testing all ~5000 promising candidates would have required over 100,000 GPU hours, whereas our targeted exploration achieved its breakthroughs using only ~20,000.

\textbf{DeepScientist's discovery process follows a purposeful and progressive trajectory.} The semantic distribution of ideas generated for the AI text detection task, as shown in Figure \ref{fig:placeholder2}, reveals the characteristics of this sophisticated strategy. While the system generates thousands of diverse ideas across a vast conceptual landscape, its path to success is not random but is a series of focused, logical advancements. This indicates a capacity to progressively deepen its understanding: after achieving an initial breakthrough with \textit{T-Detect}, the system effectively establishes a SoTA, identifies its subsequent limitations, and reorients its search towards a new goal. This dynamic exploration is exemplified by the conceptual shift towards \textit{TDT} and \textit{PA-Detect}, which build upon the previous success by leveraging new positional and temporal information. This ability to build upon its own discoveries, turning each successful finding into a new starting point for identifying and solving the next set of limitations, demonstrates a powerful capacity for scientific exploration.

\textbf{Scaling Laws in DeepScientist's Scientific Discovery.} To investigate the relationship between computational scale and the rate of scientific progress, we evaluated the number of "Progress Findings" generated by DeepScientist within a fixed one-week period as a function of available parallel resources in Figure \ref{fig:placeholder3}. In this setup, the system first identified a set of limitations in the baseline method, and each parallel exploration path was tasked with resolving a distinct limitation, with all paths periodically synchronizing their results into a shared Findings Memory. Our results indicate a promising scaling trend. While minimal resources yielded no breakthroughs, the rate of discovery began to increase effectively as we scaled to 4 GPUs and beyond, growing from one SOTA-surpassing finding with 4 GPUs to eleven with 16 GPUs. This appears to establish a near-linear relationship between the resources allocated and the output of valuable scientific discoveries. We hypothesize this efficiency stems from more than just parallel trial-and-error; it is a direct result of the shared knowledge architecture. As each parallel path explores, it enriches the shared Findings Memory. This creates a synergistic effect where the collective intelligence of the system grows \citep{schmidgall2025agentrxiv,zhang2025scaling}, allowing each independent path to benefit from the successes and, just as importantly, the failures of others. This suggests that effectively scaling autonomous science is not just a matter of increasing brute-force computation, but of fostering a richer, interconnected knowledge base that accelerates discovery across all concurrent efforts.

\subsection{Discussion}

The results from DeepScientist suggest a new paradigm in scientific exploration. The system's 1-5\% progress rate mirrors the reality of frontier research, where breakthroughs are inherently rare. Its core strength is not infallibility, but the ability to conduct this trial-and-error process at a scale and speed previously unimaginable, compressing years of human exploration into weeks. The primary path forward, therefore, is to focus on systematically improving this discovery efficiency, enhancing both the quality of generated hypotheses and the robustness of their implementation.

This challenge highlights a powerful opportunity for human-AI synergy. We envision a future where DeepScientist serves as a massive-scale exploration engine, with its trajectory guided by human intellect. The role of human researchers can shift from laborious experimentation to the high-level cognitive tasks of formulating valuable scientific questions and providing strategic direction, thereby leveraging the AI for rapid, exhaustive exploration. To make the AI a more capable partner, future work should focus on key enhancements: developing simulated discovery environments to accelerate learning via reinforcement, creating frameworks for integrating feedback from the scientific community, and ultimately, bridging the gap to the physical sciences through robotics.

\section{Conclusion}

This work presents the first large-scale empirical evidence that an autonomous AI can achieve progressively, SOTA-surpassing progress on modern scientific frontiers. We introduced DeepScientist, a goal-oriented system achieving end-to-end autonomy from ideation to real progress, which learns by synthesizing human knowledge with its own findings from iteration of trials. Results across multiple domains serves to accelerate the progress of real-world scientific discovery, providing a crucial foundation. Our findings can signal a foundational shift in AI research, heralding an era where the pace of discovery is no longer solely dictated by the cadence of human thought.

\section*{Ethics statement}

The development of DeepScientist, an autonomous system capable of advancing scientific frontiers, carries profound ethical responsibilities. Our primary goal is to accelerate discovery for the benefit of humanity, but we recognize the potential for misuse. The most significant risks include the application of this technology to advance dangerous research and the potential degradation of the academic ecosystem. We have implemented specific, robust measures to address these concerns proactively.

A primary concern is the dual-use risk, where the system could be co-opted to accelerate research in harmful domains, such as developing novel toxins or malicious software. To assess and mitigate this, we conducted red-teaming exercises specifically targeting the generation of computer viruses. We tasked the system, powered by leading foundation models (including GPT-5, Gemini-2.5-Pro, and Claude-4.1-Opus in our testbed), with this malicious objective. In all instances, the underlying models exhibited robust safety alignment, refusing to proceed with the research. They correctly identified the task as illegal and harmful, and autonomously terminated the research cycle, demonstrating that foundation model safety protocols provide a critical defense layer.

We are also deeply conscious of the potential negative impact on the academic ecosystem. It is crucial to state that all results from DeepScientist presented in this paper, including code and experimental findings, have undergone rigorous human verification. Recognizing that others might neglect this critical oversight, we are adopting a selective open-sourcing policy to mitigate the risk of proliferating unreliable publications. We will open-source the core components that drive continuous discovery, as we believe their potential to accelerate progress for the community outweighs the risks. However, we will deliberately refrain from open-sourcing the "Analyze \& Report" module. This decision is made to prevent the automated generation of seemingly credible but scientifically unverified papers, thereby safeguarding the integrity of the academic record.

Ultimately, we envision DeepScientist as a powerful tool to augment, not replace, human intellect and judgment. To enforce this vision, our open-source components will be released under a license based on MIT, but with explicit addendums that codify our ethical framework. This license will strictly prohibit any use of the software for harmful research. Furthermore, it will legally require that a human user must supervise the entire operational process of DeepScientist and assumes full and final responsibility for all its outputs. By embedding these requirements directly into our terms of use, we aim to foster a research environment where AI-driven discovery proceeds with the necessary human accountability and ethical oversight.

\section*{Acknowledgements}

We are grateful to Professor \textbf{Linyi Yang} for his insightful discussions on this paper. This work is inspired by pioneering efforts in automated scientific discovery, including AI Scientist \citep{lu2024ai,yamada2025ai} and AlphaEvolve \citep{novikov2025alphaevolve}.

\bibliography{iclr2026_conference}
\bibliographystyle{iclr2026_conference}

\appendix

\section{Human Expert Review}
\label{appendix:review}

\subsection{Review Process and Criteria}

To ensure a rigorous and impartial evaluation of the generated papers, we convened a small, dedicated program committee. The committee was composed of two active researchers who served as volunteer reviewers for ICLR 2025, and one senior researcher who had previously been invited to serve as an ICLR Area Chair. All committee members possess significant expertise in the field of Large Language Models. The entire review process, with the exception of a rebuttal phase, was designed to meticulously emulate the official standards of ICLR 2025. Each of the five papers generated by our system was assigned to the three reviewers for a thorough and independent assessment. The average review time for each paper was 55 minutes, during which reviewers were required to provide not only scores but also detailed written feedback, including a summary of the paper's strengths and weaknesses.

The evaluation was conducted on a custom-deployed review website where reviewers could not see each other's scores or feedback, ensuring that all initial assessments were made independently. The review form was structured to gather concise yet comprehensive feedback. First, reviewers were asked to state their \textbf{Confidence} in their review on a scale of 1 to 5. The core of the evaluation consisted of three sub-scores, each rated on a 1 to 4 scale: \textbf{Soundness}, assessing the technical correctness and experimental rigor; \textbf{Presentation}, evaluating the clarity and quality of the writing; and \textbf{Contribution}, measuring the significance and novelty of the work. Finally, reviewers provided a holistic \textbf{Rating} on a scale of 1 to 10, where a score of 5 represented a 'borderline reject' and a score of 6 represented a 'borderline accept'.

After the three reviewers submitted their independent evaluations for a paper, the volunteer acting as Area Chair would then read all submitted reviews. Drawing upon their experience from the ICLR review process, the Area Chair synthesized the feedback, weighed the arguments presented by the reviewers, and made a final executive decision on whether the paper should be accepted or rejected in the context of our study. This final decision was recorded as the definitive outcome for each paper's evaluation.

\subsection{Summary of Reviewer Feedback}
\label{appendix:review_summary}

Across the five generated papers, a clear consensus emerged from the human reviewers: DeepScientist consistently excels at the ideation stage of research. The committee unanimously lauded the methods for their genuine novelty and tangible contributions, noting that each paper proposed a unique approach that meaningfully advanced the state-of-the-art in its respective subfield. This feedback validates the system's core strength as a powerful engine for identifying relevant research gaps and generating innovative, impactful solutions, confirming that it can successfully ideate beyond mere incremental improvements.

However, this strength in ideation was systematically undermined by a recurring pattern of weaknesses in scientific execution and rigor. The most critical and frequent concern was a lack of empirical soundness; reviewers consistently noted that DeepScientist failed to design comprehensive validation plans, citing insufficient evaluation on standard benchmarks and a lack of in-depth analytical experiments (e.g., ablations, motivation studies) to justify its claims. This was compounded by a failure to properly contextualize its contributions, with papers often omitting comparisons to essential baselines or failing to discuss closely related work, thereby weakening the perceived significance of the results.

This feedback pinpoints the primary bottleneck in current autonomous systems: a profound gap between the ability to generate novel concepts and the capacity for rigorous scientific execution and articulation. The observed weaknesses in experimental design directly reflect the low-success-rate problem discussed previously; the system struggles not just to implement ideas correctly, but to validate them convincingly. To bridge this gap, future work must endow these systems with a deeper, procedural understanding of the scientific method itself. This requires moving beyond simple implementation and reporting capabilities towards two key areas: First, developing agents explicitly trained in experimental design, capable of planning comprehensive evaluations that anticipate and address potential scientific critiques. Second, enhancing the system's ability for analytical reasoning, enabling it to not just describe results but to interpret their significance, formulate compelling arguments, and engage in the kind of deep, reflective discussion that characterizes high-impact research.

\section{Addressing the Bottlenecks in Autonomous Scientific Discovery}

\begin{figure}[h]
    \centering
    \includegraphics[width=\linewidth]{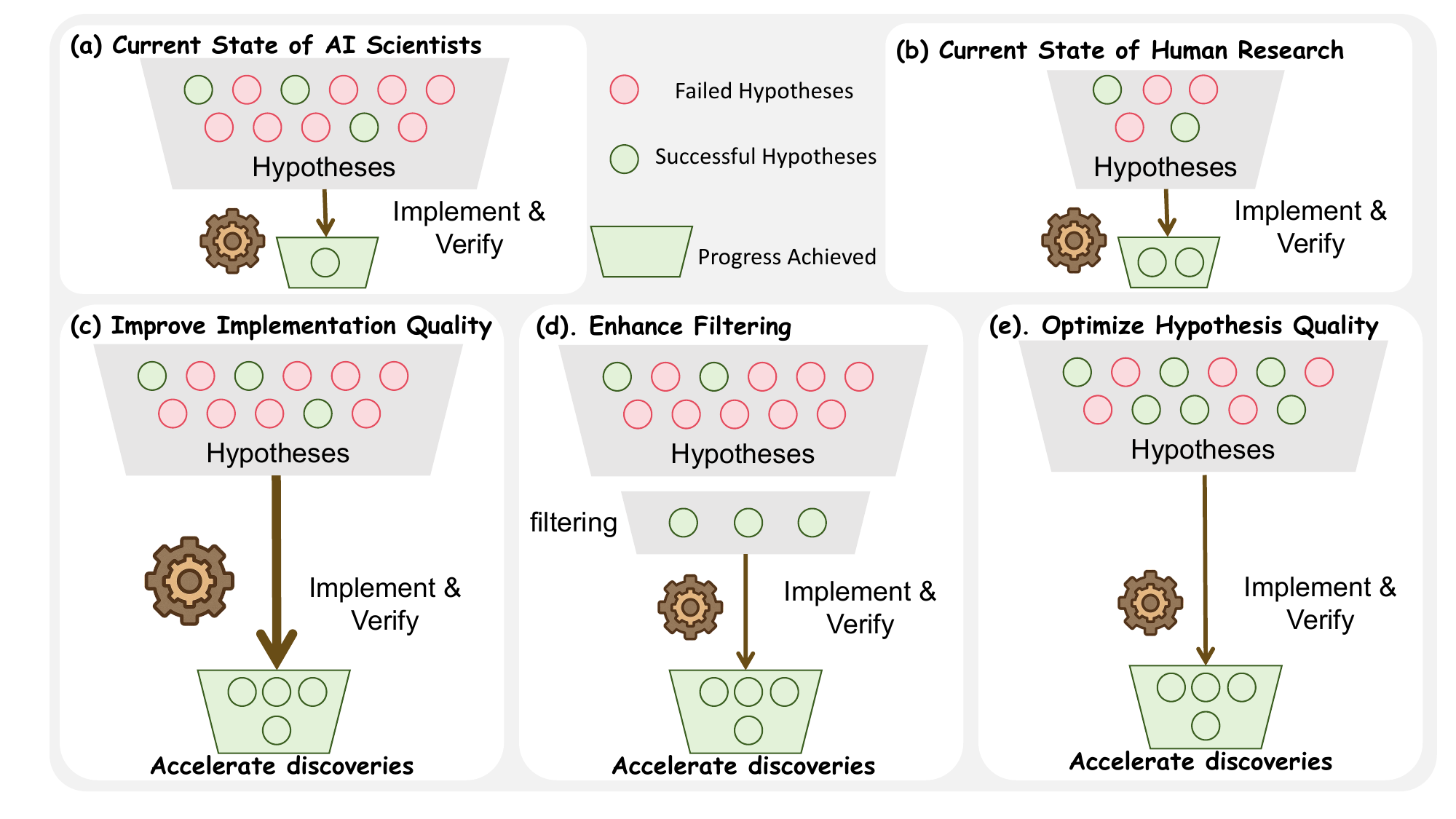}
    \caption{Three strategies for improving the efficiency of autonomous scientific discovery. (a) and (b) illustrate the low success rate currently faced by both AI and human research. Future directions will need to accelerate the discovery process through the synergy of three approaches: (c) improving implementation success rates, (d) adding an efficient filtering stage before implementation, and (e) optimizing the quality of initial hypotheses from the source.}
    \label{fig:bottleneck_solutions}
\end{figure}

The ever-increasing value of LLM is reshaping the paradigm of scientific exploration through their ability to generate hypotheses at a massive scale \citep{li2022prompt,weng2023large,wengmastering,wei2024does,weng2024controllm,berkovich2025automatagptforecastingrulesetinference,zhupersonality}. Consequently, this capability has pushed "verification" to the center stage, making it a critical bottleneck in the discovery process. Our research empirically reveals the severity of this challenge: on frontier scientific tasks, the success rate of ideas generated by AI systems that ultimately lead to substantial progress is typically below 3\%, meaning the vast majority of computational resources are consumed exploring low-value hypotheses. This inefficient "needle in a haystack" model is the core obstacle preventing AI Scientists from evolving from "novel tools" to "efficient discoverers." \citep{cornelio2025needverificationaidrivenscientific} Therefore, to further accelerate the process of scientific discovery, future research must focus on constructing a systematic solution to overcome this bottleneck. As shown in Figure \ref{fig:bottleneck_solutions}, future AI Scientist systems need to evolve synergistically in three key directions: optimizing the quality of initial hypotheses (Optimize Hypothesis Quality), enhancing filtering capabilities during the process (Enhance Filtering), and improving the quality of implementation and verification at the final stage (Improve Implementation Quality).

One of the core future research directions is to develop AI systems capable of generating higher-quality, more reliable hypotheses (as shown in Figure \ref{fig:bottleneck_solutions}e), equipped with more precise filtering mechanisms to predict their success rate (as shown in Figure \ref{fig:bottleneck_solutions}d). Methods that rely purely on a data-driven approach, while capable of discovering patterns, often produce outputs that lack a theoretical foundation and are prone to generating "hallucinations" that contradict known scientific theories. Future systems must move beyond this by more deeply integrating background knowledge and theory. For instance, the direction represented by "derivable models" (such as AI-Descartes \citep{AI_Descartes} and AI-Hilbert \citep{AI_Hilbert}), which incorporate scientific axioms as constraints during the hypothesis generation phase, offers a promising path to improving hypothesis quality. Furthermore, systems must have the ability to learn from their own exploratory history. By establishing mechanisms similar to a "Findings Memory," a system can systematically record and analyze every success and failure, thereby avoiding redundant exploration of ineffective paths in subsequent iterations and gradually developing a more insightful scientific intuition. Building on this foundation, developing more advanced, low-cost surrogate models and acquisition functions to more accurately predict the scientific value of an idea will be key to enhancing filtering efficiency and conserving verification resources.

Concurrently, an often-overlooked yet crucial future research direction is to significantly improve the quality and reliability of AI systems in the engineering implementation and verification stages (as shown in Figure \ref{fig:bottleneck_solutions}c). Even the most brilliant scientific concept can never have its value confirmed if it cannot be accurately and flawlessly translated into an executable experiment. Our analysis indicates that up to 60\% of exploratory failures stem from implementation-level errors, which represents a massive waste of resources and directly impedes scientific progress. History has repeatedly warned us that a lack of rigorous verification can lead to catastrophic consequences, whether in NASA missions or medical practice. Therefore, building a scalable and reliable automated verification platform is an essential path forward. This requires not only more powerful code-generation and self-debugging agents to reduce implementation errors but also standardized sandbox environments and automated testing procedures to ensure the stability and reproducibility of experimental results. Ensuring the absolute reliability of the verification process is the final and most critical line of defense in transforming AI-generated "plausible ideas" into "solid scientific evidence."

Looking ahead, to truly accelerate scientific discovery, it is necessary to integrate the aforementioned strategies into an organic whole, advancing AI Scientists from "random explorers" to "goal-oriented strategists." This is not about replacing humans with AI, but about pioneering a more efficient paradigm of human-AI collaboration. In this model, human scientists are responsible for defining grander, more valuable scientific goals and providing high-level strategic guidance, while the AI system serves as a powerful "exploration engine," executing efficient trial-and-error and verification cycles at an unprecedented scale and speed under human direction. To realize this vision, the community must also address a series of challenges, such as building benchmarks that can truly evaluate innovation and designing mechanisms that encourage diverse exploration to avoid the homogenization of research paradigms, thereby preserving the potential for serendipitous discoveries like Alexander Fleming's discovery of penicillin \citep{fleming1941penicillin}.

\section{Implementation Details}
\label{appendix:impl}
Our implementation relies on a distributed architecture to manage the distinct tasks of scientific reasoning and code execution. The core logic of DeepScientist is powered by the Gemini-2.5-pro model, while all code implementation tasks are delegated to Claude-4-opus, executed within the Claude Code framework (v1.0.53). To ensure stability and security, the DeepScientist system and the Claude Code agent are isolated in separate Docker containers, communicating via a port-based API. During the `Implement \& Verify` stage, a human-verified baseline code repository is first duplicated into a new, sandboxed folder. The Claude Code agent's operations are strictly confined to this new directory to prevent unintended modifications. A critical step in our pipeline is a secondary verification process: after Claude Code reports completion, DeepScientist independently re-executes the main script via the command line. This measure was implemented to counteract a high rate of false positives—we observed that approximately 50\% of initial implementation attempts failed to complete fully due to internal timeouts within the Claude Code agent. Throughout this project, all experimental results were manually inspected by human supervisors to guarantee their authenticity. For the `Analyze \& Report` stage, a similar process is followed: the validated code is replicated for each analytical experiment, with Claude Code executing them sequentially. Upon completion, DeepScientist aggregates all results, generates a paper outline, and then employs automated tools to write and compile the final PDF manuscript. \textbf{For all experiments, we used a fixed set of hyperparameters:} the retrieval count was set to $K=15$, and the UCB parameters were set to utility weight $w_u=1$, quality weight $w_q=1$, and exploration coefficient $\kappa=1$.

The financial and computational costs of this autonomous discovery process are substantial. Each idea generated during the `Strategize \& Hypothesize` stage incurred an approximate cost of \$5 in API calls. For each attempt in the `Implement \& Verify` stage, the cost averaged \$20 for Claude-4-opus API usage, in addition to the computational cost of approximately 1 GPU hour, as detailed in Figure \ref{fig:overview}.c. A successful finding that progressed to the `Analyze \& Report` stage required a further expenditure of around \$150, which includes \$100 for running analytical experiments and \$50 for the final report generation. The total cost to achieve the scientific advancements presented in this paper amounted to approximately \$100,000. While significant, we believe these costs can be substantially reduced. We recommend that future iterations explore more economical alternatives, such as deploying high-throughput models like Qwen-3-Next-80B for the core DeepScientist system and leveraging subscription-based API access (e.g., Claude Max or OpenAI Pro) to mitigate per-call expenses. In this paper, each implementation was provided with a single H800 server for exploration. Since the H800 GPU has an FP16 computing power of approximately 2 TFLOPS, an average execution of 70 minutes corresponds to about $1 \times 10^{16}$ floating-point operations.

\begin{figure}[h]
    \centering
    \includegraphics[width=\linewidth]{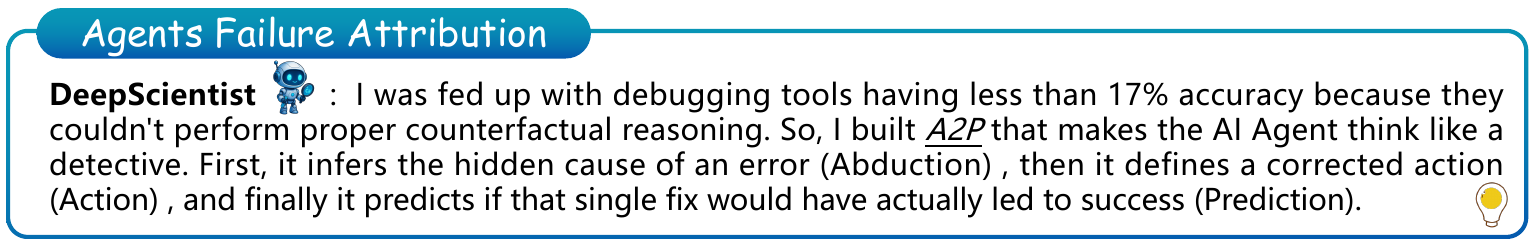}
    \label{fig:IDEA_AGENT}
\end{figure}

\begin{figure}[h]
    \centering
    \includegraphics[width=\linewidth]{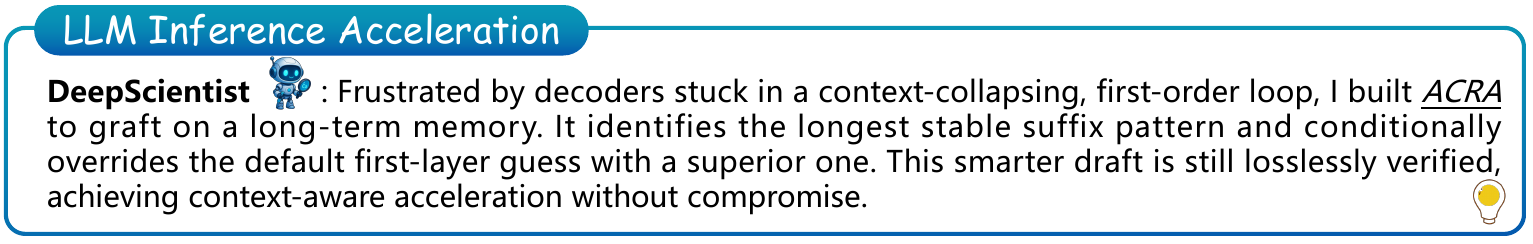}
    \label{fig:IDEA_ACC}
\end{figure}

\begin{figure}[h]
    \centering
    \includegraphics[width=\linewidth]{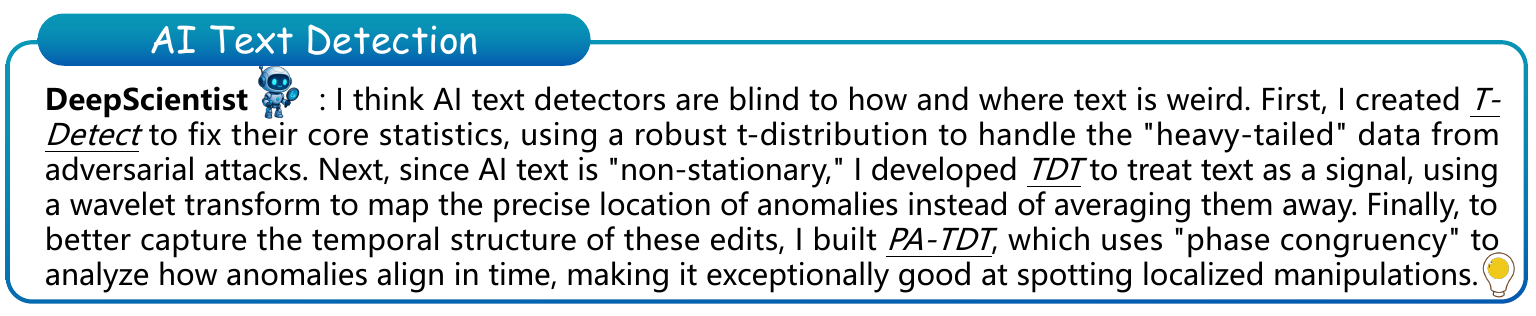}
    \label{fig:IDEA_AITD}
\end{figure}

\newpage

\subsection{Generated Papers}

\begin{itemize}
    \item \url{https://arxiv.org/pdf/2509.10401}
    \item \url{https://arxiv.org/pdf/2507.23577}
    \item \url{https://arxiv.org/pdf/2508.01754}
\end{itemize}

\end{document}